\documentclass{amia}
\usepackage{graphicx}
\pdfoutput=1
\usepackage[labelfont=bf]{caption}
\usepackage[superscript,nomove]{cite}
\usepackage{color}
\usepackage{algorithm}
\usepackage{algorithmic}

\usepackage{hyperref}
\usepackage{url}
\usepackage{subfigure}
\usepackage{booktabs}
\usepackage{multirow}
\usepackage{amsfonts}

\newcommand{\argmin}{arg\,min}

\newcommand{\Semigran}{\textbf{Semigran}}
\newcommand{\EHRTEST}{\textbf{EHR test}}
\newcommand\TODO[1]{\textcolor{red}{#1}}
\newcommand{\cut}[1]{}
\begin{document}

\title{The accuracy vs. coverage trade-off in patient-facing diagnosis models}


\author{Anitha Kannan, PhD$^{1}$, Jason Alan Fries, PhD$^{2}$, Eric Kramer, MD$^{1}$,\\ Jen Jen Chen, MD$^{1}$, Nigam Shah, MBBS, PhD$^{2}$  and Xavier Amatriain, PhD$^{1}$}
\institutes{
    $^1$Curai, Palo Alto, CA, USA $^2$Stanford University, Palo Alto, CA, USA\\
}

\maketitle

\noindent{\bf Abstract}
\textit{A third of adults in America use the Internet to diagnose medical concerns, and online symptom checkers are increasingly part of this process. These tools are powered by diagnosis models similar to clinical decision support systems, with the primary difference being the coverage of symptoms and diagnoses. To be useful to patients and physicians, these models must have high accuracy while covering a meaningful space of symptoms and diagnoses. To the best of our knowledge, this paper is the first in studying the trade-off between the coverage of the model and its performance for diagnosis. To this end, we learn diagnosis models with different coverage from EHR data. We find a 1\% drop in top-3 accuracy for every 10 diseases added to the coverage. We also observe that complexity for these models does not affect performance, with linear models performing as well as neural networks.}

\section*{Introduction}
As of {2019}, 
over one-third of adults in the USA turn to the Internet to find an answer to their medical concerns \cite{nbc19}. While the majority of online access to care is through popular search engines -- 1\% of Google search queries have a medical symptom in them \cite{hsp19} -- online symptom checkers offer patients  more directed information for guiding medical decision-making. In these online tools, patients input their initial symptoms and then proceed to answer a series of questions that the system deems relevant to those symptoms. The output of these online tools is a differential diagnosis (ranked list of diseases) that helps educate patients on possible relevant health conditions.
Online symptom checkers are powered by underlying diagnosis models or engines similar to those used for advising physicians in ``clinical decision support tools"; the main difference in this scenario being that the resulting differential diagnosis is not directly shared with the patient, but rather used by a physician for professional evaluation.

Diagnosis models must have high accuracy while covering a large space of symptoms and diseases to be useful to patients and physicians. Accuracy is critically important, as incorrect diagnoses can give patients unnecessary cause for concern. However, inferring clinical diagnoses in user-directed, online settings is non-trivial. User-reported symptoms can be vague and under-specified, creating challenges for capturing and disambiguating fine-grained symptoms.  
Using machine learning to map complex sets of symptoms to diseases entails a fundamental trade off between coverage and accuracy: in order to understand as much as possible, we would want to increase coverage, but increasing coverage may introduce more neighboring symptoms and conditions, affecting accuracy. 

To the best of our knowledge, this paper is the first in studying a central aspect of any automated diagnosis model: the trade-off between the number of diseases covered by the model and its performance for diagnosis. 
Intuitively, it is easier to train a high accuracy model for discriminating between a small number of possible diseases. However, this setting is misaligned with practice, where physicians are trained to diagnose a variety of common diseases while still being able to recognize a wide range of rarer diseases and clinical presentations that require urgent evaluation. Diagnosis models need to emulate a physician's broad diagnostic reasoning so that accuracy does not come at a high cost of being lower utility to a patient.

To study this trade-off, we need a scalable way to learn models for diagnosis with expanding disease coverage, which requires a large collection of clinical vignettes. Clinical vignettes are descriptions of patient cases that are often used for medical education.  For our purposes, clinical vignettes include symptoms, clinically significant findings (history of diabetes), and diagnosis.  We obtained our clinical vignettes from electronic health records (EHRs), which provide a more efficient and scalable approach than using curated clinical vignettes as in Razzaki {\it et al.}~\cite{razzaki18} Extracting good clinical vignettes from EHRs is challenging due to heterogeneous structure of EHR data.  Diagnosis labels on EHR data are often based on ICD codes which are primarily used for billing.  Substituting single billing codes for diagnosis as ground truth diseases can be misleading as multiple codes may represent the same presentation or group of symptoms for a diagnosis.  Vice versa, the findings associated with one particular code can be very different from the findings associated with a different code, even if they are coding for the same base diagnosis. For example, while `pneumonia" and ``tuberculosis pneumonia" are both related to pneumonia, they can have very different associated symptoms. 

The aim of this paper contrasts with previous works \cite{semigran_comparison_2016, razzaki18} where the emphasis was on benchmarking symptom checkers. These papers reviewed the performance of several symptom checkers but did not explore the impact and trade-offs associated with the number of disease and symptom targets in the benchmark dataset. 
While Razzaki {\it et al.}~\cite{razzaki18} mentions using a probabilistic graphical model that was \emph{manually} tuned using manually curated clinical vignettes, there was little clarity on which diseases and symptoms were modeled.

To summarize, the main contributions of our work include:
\begin{itemize}
    \item Introducing and analyzing the importance of the trade-off between the number of disorders and the accuracy of diagnosis models in user-facing settings. Experimental results show that an increase in 10 diseases considered can cause an approximately 1\% drop in top-3 accuracy.
    \item Introducing a methodology for building scalable patient-facing symptom checkers from electronic health records.
    \item In our problem formulation, increasing model complexity had little impact: linear models performed as well as overparameterized neural networks.
\end{itemize}

\section*{Related Work}

\noindent\textbf{Modeling differential diagnosis:} Early models for diagnosis were doctor-facing AI expert system models ({\it c.f.} Mycin\cite{mycin}, Internist-1 \cite{miller_internist-1_1982}, DXplain \cite{barnett_dxplain:_1987} and QMR \cite{rassinoux_modeling_1996}) with the goal to emulate physicians' medical diagnostic ability.  These systems have two components: an expert-curated knowledge base and an inference engine that is manually optimized. The inputs to these systems are clinical findings, which are used to derive a differential diagnosis based on the inference engine acting on the knowledge base \cite{miller_diagnostic_2016}. These models were later interpreted within a probabilistic formulation to model uncertainty in differential diagnosis \cite{ExpertSystemsProbabilities}.  In this formulation, diseases and their presentations form a bipartite probabilistic graph with the conditional probabilities and priors derived from the variables in the expert system. However, exactly computing the posterior distribution over the diseases based on the observations (input findings) is intractable in these models, and so the emphasis of this line of work has been on approximate probabilistic inference, including Markov chain Monte Carlo stochastic sampling \cite{ShweCooper90}, variational inference \cite{JaakolaJordan99} and/or  learning a recognition network \cite{Morris01} by exploiting various properties (e.g. sparsity) of the instantiated graph. 

While our paper is also focused on generating a differential diagnosis based on an input set of findings, we differ in two ways: first,  we want to use the resulting model in a patient-facing scenario (e.g. online symptom checker) where the input to the system is constrained to a set of findings gathered directly from a non-expert (non-physician) user. Second, we want to investigate the challenges in learning a diagnosis model based on clinical cases from EHR data, as opposed to manually curating a diagnosis model.

\noindent\textbf{Machine learned models from EHR}
The increasing availability of EHRs has led to surge of research in learning diagnostic models from patient data. In Wang {\it et al.}~\cite{Wang_LR_RiskPrediction}, a multilinear sparse logistic regression model is used for disease onset prediction. In Rotmensch  {\it et al.}~\cite{rotmensch_learning_2017}, a probabilistic noisy-OR model was learned to infer a knowledge graph from electronic health records, which can be used for diagnosis, similar to probabilistic inference in expert system. More recently, emphasis has been placed on directly learning diagnostic codes (ICD) predictions using deep neural networks, either instantaneously or through time (\emph{c.f.} \cite{miotto_deep_2016, Ling_DeepRL_Diagnostic, Shickel_Deep_EHR, Rajkomar18, liang2019} and references therein).  

In Rajkomar {\it et al.}~\cite{Rajkomar18}, a machine learned model was proposed to predict ICD codes from both structured and clinical notes in EHR. Similarly, Liang {\it et al.}~\cite{liang2019} introduced a model for predicting ICD codes for pediatric diseases with the provision that appropriate organ systems most relevant for the clinical case are manually designated for training.

Our work differs from these studies in two ways: first, we are interested in models for diagnosis that output the pertinent diagnosis in a manner that is useful to the patient, where it corresponds to a disease as a whole instead of only to an ICD code. For example, `acute otitis media', or the common ear infection is more useful as a grouping of ICD codes (H65.1, H66.0, H65.0, H66.0) in a patient-facing setting than each ICD code alone.  In the latter case, patients do not need to understand the difference between suppurative vs non-suppurative, they just would know they have an ear infection. Second, we would like the model to take as inputs the symptoms reported by the patients. Note that none of these approaches have been applied to the online telehealth setting where the patient using the system have no lab results or have a physical exam done.

In another recent work~\cite{ravuri18}, there were early results to show how knowledge from clinical expert systems can be combined with data from EHR to learn models for diagnosis. Similar to our setup, this paper also used diagnosis labels. However, the experiments that involved EHR was done on a small-scale corresponding to seven easily discernable diseases from clinical notes of patients admitted in the hospital. In contrast, we are interested in understanding the trade-off between accuracy and coverage. 


\noindent\textbf{Evaluation of symptom checkers:} 
In a recent study in Semigran {\it et al.}~\cite{semigran_comparison_2016}, diagnosis and triage
accuracy of over 50 competing symptom checkers are evaluated. 
This provides a useful benchmark study for use for subsequent studies.
Since then, there have been subsequent studies comparing the accuracy of the
symptom checkers against the gold-standard
performance of human doctors on the same test set \cite{razzaki18}. 

In Razzaki~{\it et al.}~\cite{razzaki18}, a hand-coded probabilistic model (explicit details of the model are not provided in the paper) is manually tuned. In particular, an iterative procedure is used to \emph{manually} tune the parameters of the graphical model, based on curated clinical cases.  This can potentially explain the high accuracy on the test set especially since the details on the setup are not provided, such as how broad their coverage was of the diseases or findings in their underlying model. For our paper, we use the same evaluation set as that provided in Semigran {\it et al.}~\cite{Semigran} but do not use any manually curated clinical cases or perform manual tuning; instead we train machine learning models for diagnosis with the goal to understand the trade-off and fidelity of these models.


\section*{Methods}
\subsection*{From electronic health records to medical diagnosis training data}
Electronic health records (EHRs) provide access to large sets of patient-level clinical information, including patients symptoms and their corresponding diagnosis. This data can be useful for training models for medical diagnosis; however, this data is also extremely noisy. For instance, patients can present with complex medical histories and co-morbid diseases yielding a constellation of findings, all of which can be associated with multiple diagnoses. Moreover, patients have several visits over time, complicating the demarcation between the start and end of a disease.

We use electronic phenotyping \cite{Banda18} to identify patients with diseases of interest. 
The phenotypes that defines a disease is assumed to be known. Algorithm~\ref{algo:ehr} outlines our approach to phenotyping and clinical case generation. 





\begin{figure}[H]
\centering
\includegraphics[scale=0.35]{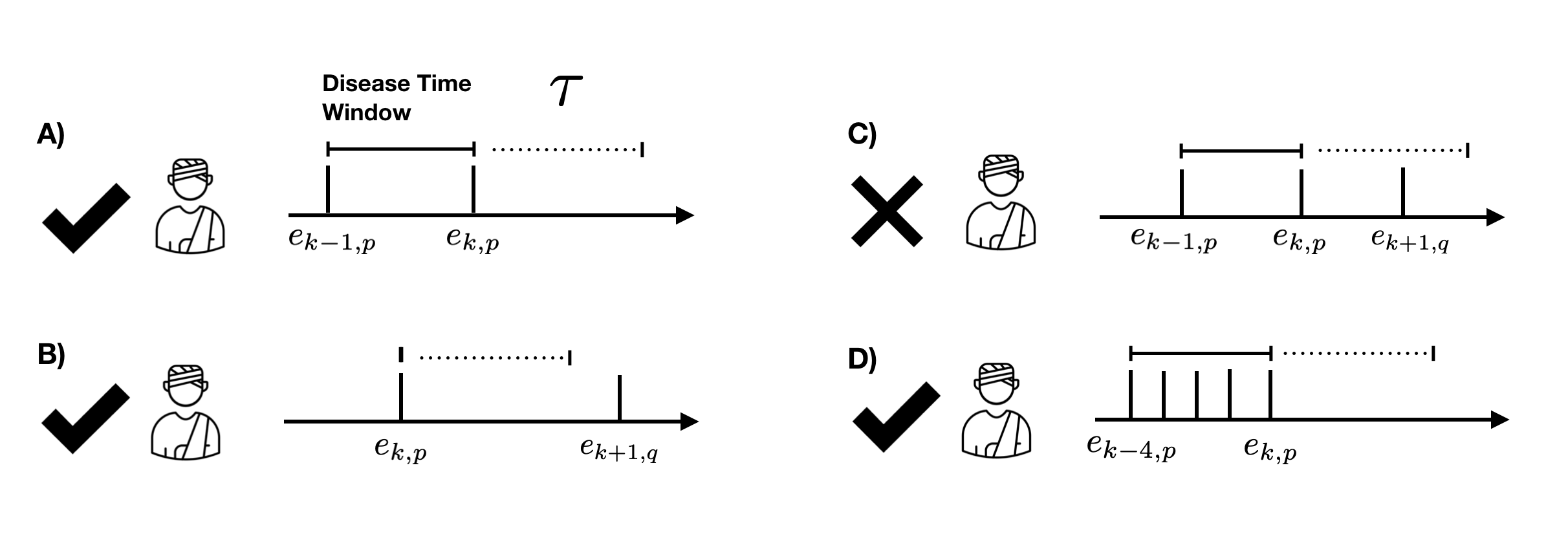}
\caption{Examples of four different patient scenarios for creating model training data from electronic health records. The disease time window is defined by the set of sequential encounters, $e_i,p$, corresponding to the phenotype of interest, $p$. Scenario C is discarded as an encounter because a different phenotype, $q$, is present within resolution period, $\tau$}
\label{fig:training_data}
\end{figure}

The ``phenotype" of disease $d$ ($\textrm{phenotype}(d)$) is the set of codes that distinctly identify the disease state. We represent the entire EHR in temporal order where data is organized by patient and by time. Each patient may have different types of records, including encounters (visits with the doctor), medications, labs etc.  For a patient $p$, let  $\mathcal{T}_p = [e_{1,p}, \cdots e_{t,p}]$ represent their timeline (chronological ordering) of their $t,p$ encounters. 
\begin{algorithm*}
\vspace{-4pt}
\caption{Algorithm for constructing clinical cases for a single disease.}
\begin{algorithmic}
\STATE {\bf Input}: A disease $d$, its phenotypes $\textrm{phenotype}(d)$ and EHR: \{$\mathcal{T}_p$\}, resolution time $\tau$
\STATE {\bf Output}: A set of clinical cases $\mathcal{D}_d =  \{(x_1, d), \cdots ,(x_T , d )\}$ corresponding to disease $d$. 
\STATE $\mathcal{D}_d \gets \emptyset$
\STATE $\mathcal{P}$ $\leftarrow$ IDENTIFY-PATIENTS(\{$\mathcal{T}_p$\}, $\textrm{phenotype}(d)$)
\COMMENT{Select patients that have diagnosis d.}
\FOR{$p \in \mathcal{P}$} 
\STATE $(t_{s,p}, t_{e,p})$ $\leftarrow$ RESOLVED-DISEASE-TIME-WINDOW(p, $\mathcal{T}_p$,  $\textrm{phenotype}(d)$, $\tau$)
\STATE $x$ $\leftarrow$ EXTRACT-FINDINGS($t_{s,p}$)
\STATE $\mathcal{D}_d \gets  \mathcal{D}_d \cup (x,d)$
\ENDFOR
\end{algorithmic}
\label{algo:ehr}
\vspace{2pt}
\end{algorithm*}
The method IDENTIFY-PATIENTS in Algo.~\ref{algo:ehr} identifies all the patients in the EHR who satisfy the following two criteria: (1) the patient $p$ has an encounter $e_{k,p}$ with at least one property in $\textrm{phenotype}(d)$ satisfied and (2) $e_{k,p}$ is an out-patient visit to mimic typical conditions encountered in a telehealth setting. For a condition to be identified as resolved, we assume that the patient has no follow-up after the diagnosis, for $\tau$ days (see illustration in Figure~\ref{fig:training_data}).  The method RESOLVED-DISEASE-TIME-WINDOW identifies the temporal window of encounters for the patient $p$ that satisfies $\textrm{phenotype}(d)$ and has not followed-up in the medical system for $\tau$ days, which is shown as the ``phenotype-present" period below.  We assign all the encounters within that temporal window to that diagnosis, assuming that the last visit corresponds to resolution of the condition. Figure~\ref{fig:training_data} provides as quick overview of valid and invalid encounters for a disease phenotype.

Once the encounter window is identified, the next step is to derive from that window a training example. EHR records may contain both unstructured clinical notes as well as structured data that captures information such as chief complaint, medications, lab orders or candidate diagnoses for that encounter. The unstructured clinical note provides more detailed description of both subjective and objective components, assessment and plan, and is particularly important for capturing all the subjective content related to the patient visit. We use the clinical note corresponding to the first visit of that temporal window for the diagnosis ($e_{k-1,p}$ in the Fig.~\ref{fig:training_data}).

The first visit clinical note enables identifying the main chief complaints and history of present illness that eventually led to the diagnosis. The subsequent notes  tend to focus on other clinically relevant measurements including laboratory measurements and imaging that are not available in a patient-facing interview situation. The EXTRACT-FINDINGS method takes clinical note as the input and returns a set of patient-facing symptoms present in the note. Internal to this method is access to universe of symptoms that can be asked in a patient-facing setting. The implementation of this method involves two stages: First, an in-house machine learned model does named entity recognition to identify entities in clinical text. Next, a smart lookup based method is used to constrain the entities to those are applicable in a patient facing setting (answerable by the user).

\noindent\textbf{Constructing the training dataset}
We considered a dataset of outpatient primary care patient visits at Stanford Health Care from February 2015 to November 2016. We restricted to only patients in the age range of 18-50 years to limit the effect of co-morbidities and chronic conditions.  We generated clinical cases pertaining to each disease according to algorithm~\ref{algo:ehr} using their corresponding phenotypes. To confirm disease resolution, we use $\tau=30$ days as the duration without a follow-up. 

To represent a broad set of differential diagnoses, we included approximately 160 diseases from a wide variety physiological categories (endocrine, cardiovascular, gastrointestinal, etc). We also ensured disorders have varying urgency - some could be treated at home while others required immediate attention by a physician. While we excluded very rare diseases, the disease set generally had a wide range of prevalence. The ICD-9 and ICD-10 codes that represented each disease phenotype required manual curation by medical experts.  Although a disease could clearly be related to many ICD codes, not all of those codes reflect a typical manifestation for a disease and we chose the most generic code(s) to represent the disease. For example, the disease `pericarditis' can have several representations including $I31.9$ for `Diseases of the pericardium, unspecified', $I30$ for `Acute pericarditis' and  $I01$ for `Rheumatic fever with heart involvement'. For our use case of patient facing diagnosis, we choose only the codes that correspond to typical presentation of the disease in its acute form without comorbidities or concurrent pathologies. This would mean that we would exclude $I01$ for acute pericarditis because its underlying cause is Rheumatic fever. 

\cut{
we need to choose only the codes that is representative of the 
\begin{itemize}
\item I31.9 Diseases of the pericardium, unspecified: this is the most generic code for pericarditis but is very nonspecific and could include other pericardial disorders.
\item I30 Acute pericarditis:  this code relates to a sudden manifestation of pericarditis, separate from chronic pericarditis which is a recurring form of pericarditis.
\item I01 Rheumatic fever with heart involvement:  this is an example of how a disorder elicits another disorder - rheumatic fever is the cause of the pericarditis - which can greatly change the group of findings that this maps to.
For our disease list, we chose the codes that were most representative of a typical presentation of the disease in its acute presentation.
\end{itemize}}


The resulting dataset had 79,355 patients, covering 103,989 encounters of 164 diseases of interest. On average, there were 1.8 valid encounter windows. Each clinical case constructed had 14.75 positive and 8.1 negative findings. Negation of findings were found using NegEx \cite{negex}. 10\% of patients and their encounters were held as validation set for model tuning, and the remaining 90\% were used for model training. In the Results section, we describe the evaluation set used for model evaluation.

There is a large class imbalance in the dataset as some diseases were more prevalent than others. Therefore, we adopted the following strategy for data balancing during training: we limited the maximum number of examples per disease to be 3000 through undersampling. For diseases that have less than 3000 examples, we up-sample with replacement.


To study the trade-off between the accuracy of the diagnosis model as a function of the number of diseases, we constructed the following derived datasets:  $\mathcal{D}_0$, $\mathcal{D}_{+20}$, $\mathcal{D}_{+40}$ and $\mathcal{D}_{+60}$.
$\mathcal{D}_0$ corresponds to the dataset with labels that are of interest. In our setting, we used the diseases that are part of the \Semigran~evaluation set (discussed in evaluation dataset in Results section) as $\mathcal{D}_0$. The subsequent datasets correspond to adding training cases for additional diseases (+x) to $\mathcal{D}_0$ and 20 extra diseases over $\mathcal{D}_{+x-20}$. As example, $\mathcal{D}_{+40}$  corresponds to the dataset with 40 extra labels (diseases) over  $\mathcal{D}_0$ and 20 extra labels over $\mathcal{D}_{+20}$. The training cases for a label is same across datasets. Curated by the medical team, these additional diseases satisfy at least one of two properties: (a) it is a disease that may need to be ruled out when considering a differential diagnosis from $\mathcal{D}_0$ or (b) it is a common disease in a tele-health setting. 

\cut{
\begin{figure}[H]
\centering
\includegraphics[scale=0.35]{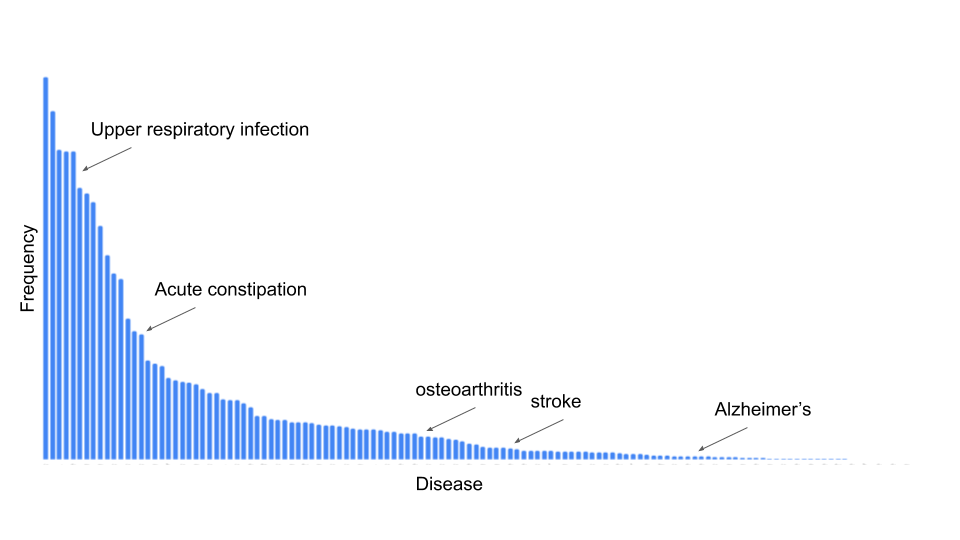}
\caption{Long tail in the disease prevalence in training set. We balance the dataset by sampling with replacement. \TODO{Left Yaxis unlabeled if this would be legal concern for Stanford}}
\label{fig:datadist}
\end{figure}
}

\label{sec:ehr}
\subsection*{Models}

We are interested in a machine learned approach that takes symptoms as input, and outputs a ranked list of diagnosis (differential diagnosis). Ideally, we can formulate this as a ranking problem, with rank-ordered training labels. However, electronic health records often provide the set of diseases in differential diagnosis but thye are not always correctly ordered. Therefore, we formulate differential diagnosis as a classification task, and use the output scores from the model to rank the diagnosis.

Let  $\mathcal{S} = \{s_1, \cdots , s_K\}$ be the universal set of $K$ symptoms that can be elicited from the patients. Let $\mathcal{Y} =  \{1, ..., L\}$ correspond to $L$ diagnoses that are of interest.

We assume access to a labeled data set of $N$ clinical cases, $\mathcal{D} = \{(\mathbf{x}_1,  y_1), ... , (\mathbf{x}_N, y_N)\}$. For each case $n$, $y_n \in \mathcal{Y} $ is corresponding diagnosis for the case. Each element $x_{n,j} \in \{-1, 0, 1\}$ represents whether the symptom $x_j \in \mathcal{S}$ is present ($1$) absent ($-1$) or unknown ($0$) in the corresponding $n^{th}$ patient. Note that the vast majority of symptoms are unknown in each clinical case, yielding a sparse representation.

The goal is to learn a function from $f: \mathcal{X} \mapsto \mathcal{Y}$ that minimizes empirical risk:
  $\min_{f \in \mathcal{F}} \sum_{n=1}^{N} \delta(y_n, f(\mathbf{x}_n))$, 

where $ \delta: \mathcal{Y} \times \mathcal{Y} \mapsto \mathbb{R}$ measures the loss incurred in predicting $ f(\mathbf{x}_n)$ when the true label is $y_n$. During evaluation we use 0/1 loss and during training we use the surrogate cross entropy loss between the delta one-hot encoding of $y_n$ and the output softmax probability vector provided by the function $f$. Note that having a binary 0/1 loss is a simplification and we would ideally have a loss function that is cost-sensitive: for instance, misdiagnosis of disease that needs urgent care can lead to bad outcomes.

We experimented with different well-studied classification models, including multiclass logistic regression and multilayer perceptron with and without distributed representation (using embedding layer) for symptoms, in order to study the hypothesis of whether a non-linear model that takes into account dependencies between symptoms are more effective for this task.

\noindent\textbf{Multiclass logistic regression:} This model learns a linear mapping between $\mathbf{z}$ and label $y$, with the mapping function defined by:
\begin{equation}
    P(y = k|\mathbf{s},W, \mathbf{b}) = \frac{\exp (b_k + \mathbf{z(\mathbf{x})}. \mathbf{w}_k)}{1 + \sum_{j} \exp(b_j+\mathbf{z(\mathbf{x}}) . \mathbf{w}_j)}
\end{equation}
where $\mathbf{w}_j$ is a vector of weights, $b_k$ is the bias for the $k^{th}$ class and  $\mathbf{z(\mathbf{x}})$ is featurized representation of inputs $\mathbf{x}$. 

In our setting, $\mathbf{z(\mathbf{x}})$  is a binary feature vector that is  $2 \times |\mathcal{S}|$ to model presence and absence of  symptoms. One can also think of using $|\mathbf{S}|$ dimensional ternary vector corresponding to presence, absence and none. In practice, we found that the clinical notes may specify the same symptom as positive or negative in different regions of the clinical note. For instance, a person coming in with no fever may be asked to follow-up when he or she has a fever. This means that in the note `fever' is mentioned in both senses; however, there is only one valid sense, in this case absent. We decided to model present and absent symptoms separately to let the machine learned model infer the correct predictor from the model. We leave as a future work to design models that can discern if the symptom is present or absent or unknown in the patient, based on the entirety of the clinical note.

\cut{
Given a training set, $\mathcal{D}$, the model is trained using stochastic gradient descent by minimizing the negative log-likelihood (including an L2 regularization term):
\begin{equation}
W^*, b^* = \argmin_{W, b} - \sum_t\log P(y^{(t)}|\mathbf{s}^{(t)},W, \mathbf{b}) + \lambda \sum_{j} |\mathbf{ w}_{j}|^2,
\end{equation}
where $\lambda = 0.01$ was chosen empirically using performance on validation set.
}

\noindent\textbf{Deep neural networks:} These models can automatically capture higher order relationships of symptom-symptom interactions through layers of interaction between variables. We experimented with two versions of the deep neural networks - the basic multi-layer perceptron (MLP) and an extension that uses embeddings to represent the symptoms in continuous space (MLP-embedding). MLP uses the same input representation as logistic regression but with an architecture of two fully connected layers with a rectified linear (ReLU) activation, as opposed to single layer of connections through $\mathbf{W}$ as in logistic regression. In the case of MLP-embedding, we learn separate embeddings for the present and absent symptoms. The architecture of the MLP-embedding model includes the embedding layer for the inputs, followed by average embedding which are then passed through two layers of fully connected layers with ReLU non-linearity. Dropout layers and L2 regularization are used for controlling model capacity.

\cut{
specified in the table where $E$ represents the embedding dimensionality.~\ref{tbl:model}.
\begin{table}
\centering
\begin{tabular}{r c c l}
\toprule
\textbf{Layer} &  \textbf{(Filters, size)} &  \textbf{Output}  \\
 \midrule
Embedding &  & $F \times E$\\
Average Embedding &  &$ 1 \times E$  \\
Dropout(p) $|$ ReLU & &  $ 1 \times E $ \\
Flatten & &  E \\
Fully connected $|$ ReLU & $E \rightarrow E/2$  & $E/2$  \\
Fully connected $|$ Softmax  & E/2 \rightarrow L & L (\# of classes)  \\
\bottomrule
\end{tabular}
\caption{MLP embedding model used in our experiments. Input is padded with zeros for missing symptoms, and not used during averaging.}
\label{tbl:model}
\vspace{-10pt}
\end{table}}
All models are trained with minibatches of size 128 using stochastic gradient descent  with initial learning rate of 0.01 and momentum 0.9. All parameters of the model are updated at each step of the optimization. The parameters, including embeddings, are initialized randomly. The hyperparameters are tuned using the validation set; however for the given model architecture, we did not find substantial difference in parameters.

\subsection*{Evaluation dataset and metrics}
\noindent\textbf{Evaluation dataset:} The experiments were evaluated on the following datasets:
\begin{itemize}
\item \Semigran:  This a dataset made available as part of study in Semigran {\it et al.}~\cite{Semigran} where over 50 online symptom checkers were evaluated.  This dataset consists of 45 standardized patient clinical vignettes, corresponding to 39 unique diseases. We used the simplified inputs provided along with the clinical vignettes, as previously used in other studies. 
\item \EHRTEST: To validate the generalization of the model to a different time period, we constructed a test set from EHR corresponding to a different time period of Feb 2017-Nov 2017. This test set is constructed following the same approach that is used for training set construction. We restricted to only those diagnosis labels as in \Semigran~to understand the difference in performance and also transfer learning capability of the trained model. This resulted in 20,771 cases for the 39 unique diseases.
\end{itemize}
We want to emphasize that while \Semigran~dataset is publicly available and there have been studies on evaluating the symptom checkers on this dataset \cite{Semigran, semigran_comparison_2016, razzaki18, Fraser18}, none of the studies discuss the underlying model or the finding/diagnosis space in which the model operates. 

\noindent\textbf{Metrics:} We are interested in a metric that is valuable in deployment contexts; in particular, to aid doctors and patients in creating a differential diagnosis so that the relevant diagnoses are considered within a small range of false positives.  For this purpose, we report top-k accuracy also known as recall@k (k $\in \{1, 3,5,10,20\}$), or \emph{sensitivity} in medical literature. 
\begin{equation}
\textrm{top-k} = \textrm{recall@k} = \frac{\sum_{t=1}^{T} \sum_{j=1}^{j=K}[\hat{y}^{(t)}[j] = y^{(t)}] }{T},
\end{equation}
where $[a=b]$ is the Iverson notation that evaluates to one only if a=b or else to zero. $\hat{y}^{(t)}[j]$ is the $j^{th}$ top class predicted from a model when evaluating test case $t$.

\section*{Results}

\noindent\textbf{Learning diagnosis model from EHR}
The goal of this initial set of experiments is to establish the applicability of the diagnosis models learned from electronic health records to patient-facing situations. 

Table~\ref{tbl:semigran_0} shows the accuracy of the models trained using  $\mathcal{D}_0$ on the evaluation datasets. The results show that we can learn a model from EHR that generalizes to new datasets. Note that in this experiment, both the training set and the evaluation set have the same diseases as labels. Also, all the three machine learned models have similar performance, and this trend continues for larger number of diseases. In our setting, linear models are as effective as neural net models, and this makes sense given that the input space is high dimensional and very sparse.

In the table we also report results from Fraser {\it et al.}\cite{Fraser18} where twenty medical experts studied each case in entirety (some cases include more information such as labs that are not available in patient facing applications) and came to consensus, showing that the top-3 accuracy is still not at 100\%, showcasing the difficulty of agreeing on diagnoses even by human experts. 

\begin{table}[h]
\centering
\begin{tabular}{c|c|c|c|c|c|c}
    \toprule
    \textbf{Dataset} & \textbf{Approach} & \textbf{top-1} & \textbf{top-3} & \textbf{top-5} & \textbf{top-10} & \textbf{top-20}\\
    \hline
    \multirow{1}{*} \EHRTEST & LR & 54.19\% (.32)&	79.03\% (.25)	& 88.16\%(.18) &	95.30\%(.10) &	98.9\%(.06) \\
                            & mlp & 56.11\% (.55)&	82.11\% (.38)	& 90.79\% (.22) &	96.99\% (.11) &	99.38\% (.02) \\
                             & mlp-Embedding & 53.16\% (.94)&	79.03\% (.57)	& 88.52\% (.42) &	96.00\% (.27) &	99.19\% (.06) \\
        \hline
    \multirow{1}{*}
     \Semigran & experts\cite{Fraser18} & 72.1\% & 84.3 \%  & - & - & -\\
    & LR & 50.67\% (1.86) &	75.11\% (1.86) &	82.22\% (1.56) &	88.45\% (.99) &	91.55\% (1.86) \\
 & mlp &48.44\% (3.65) &	69.78\% (2.53) &	77.33\% (2.89)  &	84.89\% (2.89) &	90.22\% (1.25) \\ 
 & mlp-Embedding & 48.89\% (3.51) &	69.33\% (1.85) &	76.45\% (1.99) &	87.11\% (3.97) &	92.0\% (3.71) \\
\bottomrule
\end{tabular}
\caption{Model trained on $\mathcal{D}_0$. Top-K accuracy on evaluation set. Standard deviation across 5 random initialization is provided in ()} 
\label{tbl:semigran_0}
\end{table}

\begin{table}
\centering
\begin{tabular}{c|c|c}
    \toprule
    \textbf{Disorder} & \textbf{Top positively weighted symptoms} & \textbf{Top negatively weighted symptoms}\\
        \hline
    \multirow{1}{*} Lumbar strain & low back pain, back pain, & NOT low back pain, NOT numbness, \\
        & lifting, NSAID use & antibiotics, NOT back pain \\
    \hline
    \multirow{1}{*} Meningitis & headache, neck stiffness, & NOT nystagmus, NOT neck stiffness, \\
                              & shunt, paralysis & alcohol intoxication, never smoker \\
    \hline
    \multirow{1}{*} Peptic ulcer disease & abdominal pain, epigastric pain, & rash, acute, \\
                                         & black stool, anxiety & allergy, chills\\
    \hline
    \multirow{1}{*}  URI & cough,throat pain, &  NOT fever, NOT exudate,	\\
                & 	congestion,	nasal congestion & abdominal pain, recent\\
    \hline
    \multirow{1}{*}  Acute pharyngitis & throat pain, exudate, & NOT exudate, NOT cough,\\  
                                       & swollen glands, fever & NOT fever, NOT nasal congestion \\
     \hline
    \multirow{1}{*}  UTI& 	dysuria, urinary frequency, & NOT blood in urine, NOT flank pain\\
    & sexual intercourse, antibiotics & NOT flank tenderness, NOT back pain \\ 
  \bottomrule
\end{tabular}
\caption{Symptoms with largest learned weights from the logistic regression model trained on $\mathcal{D}_0$} 
\label{tab:model_wts}
\end{table}

We also want to call out that the results from the study in Semigran {\it et al.}~\cite{Semigran} that releases this evaluation set; the average performance of the online symptom checkers is at 50\% in top-20 for \Semigran. In a recent study \cite{razzaki18}, results were provided for only 30 clinical cases. When extrapolated, assuming remaining 15 cases were wrongly diagnosed, their top-1 accuracy is at 46.6\% and top-3 and 64.67\%. None of these papers discuss the actual number of diseases and findings that are available to the model used for diagnosis, which makes it difficult for us to make a direct comparison. 

It was also interesting to note that there is no significant improvement in performance as we increase the complexity of the model. We also refer readers to the supplementary section of Rajkomar {\it et al}~\cite{Rajkomar18} where similar observation was made: a simple linear model gave most of the accuracy and ensembling with non-linear deep nets provided small improvement gains. Since our goal is to understand the performance trade-off, we did not focus on ensembling strategies.  In Table~\ref{tab:model_wts}, we present the top-4 symptoms with largest positive and negative weights learned by the logistic regression model. The model has learned to hone in on symptoms that are predictive of the underlying disease. 



\begin{figure}[H]
\centering
\includegraphics[scale=0.22]{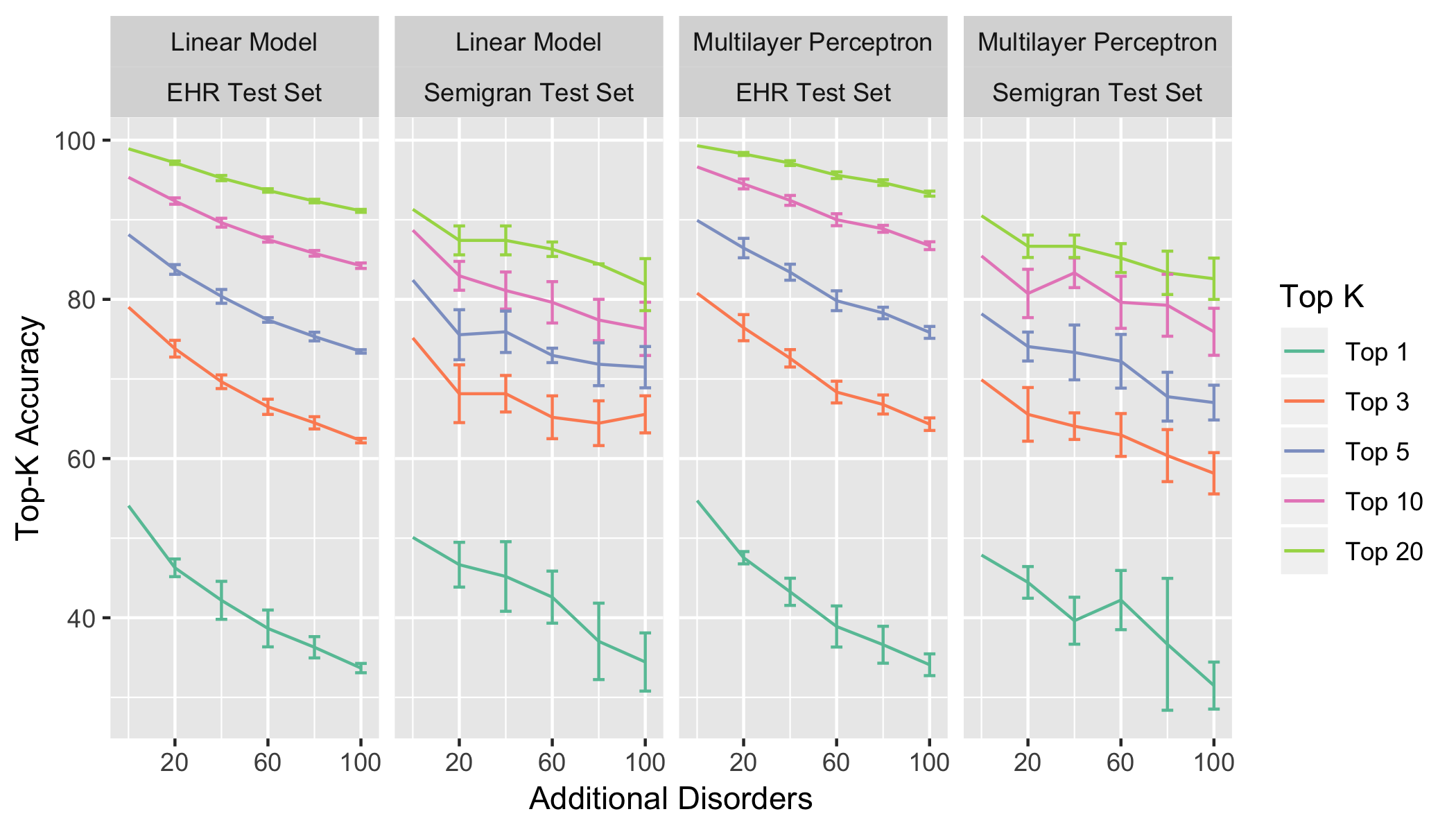}
\caption{Comparing performance of different methods on training diagnosis models with labels restricted to diseases in \Semigran. Error bars over 5 different random sampling of the additional disorders being added, while keeping the random seed fixed. We found similar trends with the neural network model that uses embedding, and do not show results due to space constraints. Error bars computed on 5 different random choices of the additional diseases added.}
\label{fig:comparison}
\end{figure}

\noindent\textbf{Diagnosis accuracy vs. disease coverage}
Figure~\ref{fig:comparison} compares the performance for models trained on different training data splits from $\mathcal{D}_0 \cdots \mathcal{D}_{+100}$. Top-k accuracy decreases for all $k$ and for both test sets as the number of additional diseases is increased in the training set. This decrease in accuracy is largest for top-1 accuracy, which drops from 49.6\% to 33\% on average, and smallest for top-20 accuracy, which drops from 90.8\% to 83.6\% on average. 

We can characterize the slope of the above lines according to $A_k = \beta_\mathcal{D} * |\mathcal{D}_k| + \beta_{M} $, where $A_k$ is the top-$k$ accuracy, $|\mathcal{D}_k|$ is the number of additional diseases added to the disease space, $\beta_\mathcal{D}$ is a coefficient for the size of the disease space and $\beta_M$ is a coefficient for the model architecture. We can see from tbl.~\ref{tbl:regression_coefficients}, that all top-k accuracies drop with increasing number of diseases as shown by negative values for $\beta_\mathcal{D}$. This drop is most significant for top-1 accuracy where the addition of every additional disease results in a drop of $0.156\%$ points in accuracy as represented by $\beta_\mathcal{D}$.

\begin{table}[ht]
\centering
\begin{tabular}{l|r|r|r|r}
    \toprule
     & $\beta_\mathcal{D}$  & \textbf{Std. Err.} & \textbf{t-value} & \textbf{p-value}\\\hline
    Top-1 & -0.156 & 0.010 & -15.52 & $\leq$ 2e-16 \\
    Top-3 & -0.104 & 0.0078 & -13.47 & $\leq$ 2e-16 \\
    Top-5 & -0.102 & 0.0071 & -14.41 & $\leq$ 2e-16\\
    Top-10 & -0.092 & 0.0074 & -12.45 & $\leq$ 2e-16 \\
    Top-20 & -0.070 & 0.0056 & -12.40 & $\leq$ 2e-16 \\ 
\bottomrule
\end{tabular}
\caption{Regression coefficients demonstrating the change in top-k accuracy for each additional disease added to the disease space. We find decreasing accuracy, represented by a negative $\beta_\mathcal{D}$, across all models }
\label{tbl:regression_coefficients}
\end{table}

These results have an important consequence. The evaluation set is fixed and corresponds to a small set of diseases that are of interest. The models are trained with increasing coverage of diseases (trained using $\mathcal{D}_0$ to $\mathcal{D}_{+100}$),  to reflect the need to increase model's ability to diagnose more diseases when deployed. However, from Figure~\ref{fig:comparison}, we can see that the performance of the models on the fixed evaluation set drops as we increase the coverage of diseases (trained from $\mathcal{D}_0$ to $\mathcal{D}_{+100}$) at a rate of 1\% for every 10 diseases for top-3 accuracy.
\label{sec:experiments}

\section*{Discussion}
The results from this paper shed light on important considerations when we build and evaluate machine learning based models for diagnosis:
\begin{itemize}
\item Machine learning models are typically tested on the data points
with the same set of labels for which the model is trained on.
There is an inherent assumption that when deployed,
the inputs can also be classified in the same label space.
This assumption is immediately invalidated in symptom checkers 
where the patient may have any of the thousands of diseases possible. When we train classification models only for diseases of interest (e.g. diseases that can be treated in a tele-health setting), they can have better performance that are not in line with the application of building patient-facing system;  moreover, these performance metrics do not hold when deployed. Hence, when classification models are evaluated, they need to specify coverage and how that coverage affects model performance. 
\item Another important take-away is with regards to model class. As discussed earlier, deep neural networks models have been successfully used in modeling data from EHR \cite{ravuri18, Rajkomar18, miotto_deep_2016}. There can be potentially other ways of representing our problem in which neural networks may outperform. In our context, where the input is very sparse binary variables, our experiments suggest linear model has sufficient power to learn the mapping function between symptoms and diseases.
\end{itemize}
\label{sec:discussion}

\section*{Conclusion}

We outlined the importance of understanding the trade-offs between the number of disorders considered and the accuracy of diagnosis models in user-facing settings. We studied this trade off using diagnosis models trained from electronic health records. In performing the research, we created a method for training set construction for acute diseases from EHR. Our results show that there is 1\% drop in top-3 accuracy for every 10 diseases added to the diagnosis engine's coverage. We also observe that in models learned from symptoms, model complexity does not matter and linear models do as well as more complex neural network models. Given the large number of diseases to model, an alternative approach is to not only model diseases of interest but simultaneously model out-of-domain or in-the-wild inputs \cite{prabhu2019}. This would allow diagnosis model to confidently predict for only small number of diseases and be explicit about its inability to predict a diagnosis when the model is unable to make confident predictions.




\bibliography{main}

\begin{thebibliography}{10}

\bibitem{nbc19}
Weaver J.
\newblock More people search for health online.
\newblock NBC News. 2019;Available from:
  \url{http://www.nbcnews.com/id/3077086/t/more-people-search-health-online/#.XVC_UpNKiMI}.

\bibitem{hsp19}
Drees J.
\newblock Google receives more than 1 billion health questions every day.
\newblock Becker's Hospital Review,. 2019;Available from:
  \url{https://www.beckershospitalreview.com/healthcare-information-technology/google-receives-more-than-1-billion-health-questions-every-day.html}.

\bibitem{razzaki18}
Razzaki S, Baker A, Perov Y, Middleton K, Baxter J, Mullarkey D, et~al.
\newblock A comparative study of artificial intelligence and human doctors for
  the purpose of triage and diagnosis.
\newblock CoRR. 2018;abs/1806.10698.

\bibitem{semigran_comparison_2016}
Semigran H.
\newblock Comparison of {physician} and {computer} {diagnostic} {accuracy}.
\newblock JAMA Internal Medicine. 2016;176.

\bibitem{mycin}
Buchanan BG, Shortliffe EH.
\newblock Rule-Based Expert Systems: {T}he {MYCIN} experiments of the Stanford
  Heuristic Programming Project.
\newblock Addison-Wesley; 1985.

\bibitem{miller_internist-1_1982}
Miller RA, Pople HE, Myers JD.
\newblock Internist-1, an experimental computer-based diagnostic consultant for
  general internal medicine.
\newblock N Engl J Med. 1982 Aug;307(8):468--476.

\bibitem{barnett_dxplain:_1987}
Barnett GO, Cimino JJ, Hupp JA, Hoffer EP.
\newblock {DXplain}: {An} {Evolving} {Diagnostic} {Decision}-{Support}
  {System}.
\newblock JAMA. 1987 Jul;258(1):67--74.

\bibitem{rassinoux_modeling_1996}
Rassinoux AM, Miller RA, Baud RH, Scherrer JR.
\newblock Modeling principles for {QMR} medical findings.
\newblock Proc AMIA Annu Fall Symp. 1996;p. 264--268.

\bibitem{miller_diagnostic_2016}
Miller RA.
\newblock Diagnostic decision support systems.
\newblock Springer; 2016.

\bibitem{ExpertSystemsProbabilities}
Lauritzen SL, Spiegelhalter DJ.
\newblock Local Computations with Probabilities on Graphical Structures and
  Their Application to Expert Systems.
\newblock Journal of the Royal Statistical Society Series B. 1988;50(2).

\bibitem{ShweCooper90}
Shawe M, Cooper GF.
\newblock An empirical analysis of likelihood-weighting simulation on a large,
  multiply-connected belief network.
\newblock Sixth Conference on Uncertainty in Artificial Intelligence. 1990;.

\bibitem{JaakolaJordan99}
Jaakkola TS, Jordan MI.
\newblock Variational Probabilistic Inference and the {QMR-DT} Network.
\newblock Journal Of Artificial Intelligence Research. 1999;.

\bibitem{Morris01}
Morris Q.
\newblock Recognition networks for approximate inference in {BN20} Networks.
\newblock Seventeenth Conference on Uncertainty in Artificial Intelligence.
  2001;.

\bibitem{Wang_LR_RiskPrediction}
Wang F, Zhang P, Qian B, Wang X, Davidson I.
\newblock Clinical risk prediction with multilinear sparse logistic regression.
\newblock Proceedings of ACM SIGKDD International Conference on Knowledge
  Discovery and Data Mining. 2014;.

\bibitem{rotmensch_learning_2017}
Rotmensch M, Halpern Y, Tlimat A, Horng S, Sontag D.
\newblock Learning a {health} {knowledge} {graph} from {electronic} {medical}
  {records}.
\newblock Scientific Reports. 2017;7(1):5994.

\bibitem{miotto_deep_2016}
Miotto R, Li L, Kidd BA, Dudle JT.
\newblock Deep {Patient}: {An} {unsupervised} {representation} to {predict} the
  {future} of {patients} from the {electronic} {health} {records}.
\newblock Scientific Report. 2016;.

\bibitem{Ling_DeepRL_Diagnostic}
Ling Y, Hasan SA, Datla V, Qadir A, Lee K, Liu J, et~al.
\newblock Diagnostic inferencing via improving clinical concept extraction with
  deep reinforcement learning: A preliminary study.
\newblock In: Proceedings of the 2017 Machine Learning for Healthcare
  Conference. MLHC '17; 2017. p. 271--285.

\bibitem{Shickel_Deep_EHR}
{Shickel} B, {Tighe} P, {Bihorac} A, {Rashidi} P.
\newblock Deep {EHR}: {A} {Survey} of {Recent} {Advances} on {Deep} {Learning}
  {Techniques} for {Electronic} {Health} {Record} ({EHR}) {Analysis}.
\newblock arXiv:170603446. 2017 Jun;abs/1706.03446.

\bibitem{Rajkomar18}
Rajkomar A, Oren E, Chen K, Dai AM, Hajaj N, Liu PJ, et~al.
\newblock Scalable and accurate deep learning for electronic health records.
\newblock CoRR. 2018;abs/1801.07860.
\newblock Available from: \url{http://arxiv.org/abs/1801.07860}.

\bibitem{liang2019}
Liang H, Tsui BY, Ni H, Valentim CC, Baxter SL, Liu G, et~al.
\newblock Evaluation and accurate diagnoses of pediatric diseases using
  artificial intelligence.
\newblock Nature medicine. 2019;.

\bibitem{ravuri18}
Ravuri M, Kannan A, Tso GJ, Amatriain X.
\newblock Learning from the experts: From expert systems to machine learned
  diagnosis models.
\newblock Machine Learning for Health Care. 2018;.

\bibitem{Semigran}
Semigran HL, Linder JA, Gidengil C, Mehrotra A.
\newblock Evaluation of symptom checkers for self diagnosis and triage: audit
  study.
\newblock BMJ. 2015;351:h3480.

\bibitem{Banda18}
Banda JM, Seneviratne M, Hernandez-Boussard T, Shah NH.
\newblock Advances in electronic phenotyping: From rule-based definitions to
  machine learning models.
\newblock Annual Review of Biomedical Data Science. 2018;1(1):53--68.

\bibitem{negex}
Chapman WW, Bridewell W, Hanbury P, Cooper GF, Buchanan BG.
\newblock A Simple Algorithm for Identifying Negated Findings and Diseases in
  Discharge Summaries.
\newblock Journal of Biomedical Informatics. 2001;34(5):301 -- 310.

\bibitem{Fraser18}
Fraser H, Coiera E, Wong D.
\newblock Safety of patient-facing digital symptom checkers.
\newblock The Lancent. 2018;392.

\bibitem{prabhu2019}
Prabhu V, Kannan A, Tso GJ, Katariya N, Chablani M, Sontag D, et~al.
\newblock Open Set Medical Diagnosis.
\newblock CoRR. 2019;abs/1910.02830.
\newblock Available from: \url{https://arxiv.org/abs/1910.02830}.

\end{thebibliography}
\end{document}